\documentclass{article}

\usepackage{arxiv}

\usepackage[T1]{fontenc}
\usepackage[utf8]{inputenc}
\usepackage{lineno}
\usepackage{graphicx}
\usepackage{booktabs}
\usepackage{tabularx}
\newcolumntype{C}{>{\centering\arraybackslash}X}
\newcolumntype{R}{>{\raggedleft\arraybackslash}X}
\newcolumntype{L}{>{\raggedright\arraybackslash}X}
\usepackage{array}
\usepackage{makecell}
\usepackage{float}
\usepackage{multirow}
\usepackage{url}

\usepackage{subfigure}

\usepackage{xcolor}
\usepackage{tkz-kiviat} 
\usetikzlibrary{arrows}

\usepackage[linesnumbered,ruled]{algorithm2e}

\title{Relearning ensemble selection based on new generated  features}

\author{ {Robert Burduk [0000-0002-3506-6611]}\\
	Department of Systems and Computer Networks, Faculty of Electronics\\
	Wrocław University of Science and Technology\\
	Wybrzeze Wyspianskiego 27, 50-370 Wrocław, Poland \\
	\texttt{robert.burduk@pwr.edu.pl} \\
	
}

\begin{document}
\maketitle

\begin{abstract}
The ensemble methods are meta-algorithms that combine several base machine learning techniques to increase the effectiveness of the classification. Many existing committees of classifiers use the classifier selection process to determine the optimal set of base classifiers. In this article, we propose the classifiers selection framework with relearning base classifiers. Additionally, we use in the proposed framework the new generated feature, which can be obtained after the relearning process. The proposed technique was compared with state-of-the-art ensemble methods using three benchmark datasets and one synthetic dataset. Four classification performance measures are used to evaluate the proposed method.
\end{abstract}




\section{Introduction}

The purpose of the supervised classification is to assign to a recognized object a predefined class label using known features of this object. Therefore, the goal of the classification system is to map the feature space of the object into the space of class labels. This goal can be fulfilled using one classification model (base classifier), or a set of base models called an ensemble, committee of classifiers or multiple classifier system. The multiple classifier system (MSC) is essentially composed of three stages: 1) generation, 2) selection, and 3) aggregation or integration. The aim of the generation phase is to create basic classification models, which are assumed to be diverse. In the selection phase, one classifier (the classifier selection) or a certain subset of classifiers is selected (the classifier selection (CS) or ensemble pruning) learned at an earlier stage. The final effect of the integration stage is the class label, which is the final decision of the ensemble of classifiers.

There are two approaches to the ensemble selection, static and dynamic~\cite{cruz2018dynamic}. An approach has also been proposed, which combines the features of static and dynamic ensemble selection~\cite{nguyen2020ensemble}. Regardless of the division criteria for the ensemble selection, none of the known algorithms uses the relearning of classification models in the selection process of classifiers. An observation that there is a lack of the ensemble selection methods with relearning has prompted to undertake the research problem, whose aim is to develop and experimentally verify the relearning ensemble selection method.
Given the above, the main objectives of this work can be summarized as follows:
    \begin{itemize}
        \item A proposal of a new relearning ensemble selection framework.
        \item A proposal of the feature generation that is used in classifier selection process. 
        \item An experimental setup to compare the proposed method with other MCS approaches using different classification performance measures.
    \end{itemize}

This paper is organized as follows: Section~\ref{sec:RelWork} presents works related to the classifiers selection. Section~\ref{sec:mehod} presents the proposed approach to the relearning ensemble selection based on new generated features. In Section \ref{sec:experimental-setup} the experiments that were carried out and the discussion of the obtained results are presented. Finally, we conclude the paper in Section~\ref{sec:Conclusions}.

\section{Related works}\label{sec:RelWork}

For about twenty years in the literature related to classification systems there has been considered the problem of using more than one base classifiers at the same time to make a decision on whether an object belongs to a class label~\cite{sagi2018ensemble}, \cite{wozniak2014survey}. During this period, multiple classifier systems were used in many practical aspects~\cite{zhang2012ensemble}, and the ensemble pruning proved to have a significant impact on the performance of recognition systems using an ensemble of classifiers. The taxonomy of the selection methods distinguishes the static and dynamic selection. The static pruning process selects one or a certain subset of base classifiers that is invariable throughout the all feature space or defined feature subspaces. In the case of the dynamic selection, knowledge about the neighborhood of the newly classified object is used (most often defined by a fixed number of nearest neighbors) to determine one or a certain subset of the base classifiers for the classification of a new object.

The discussed topic is still up-to-date, as evidenced by the propositions of new the ensemble selection methods. The dynamic programming-based ensemble pruning algorithm is proposed in~\cite{alzubi2020optimal}, where the cooperative game theory is used in the first phase of the selection and a dynamic programming approach is used in the second phase of the selection procedure. The Frienemy Indecision Region Dynamic Ensemble Selection framework~\cite{oliveira2017online} pre-selects base classifiers before applying the dynamic ensemble selection techniques. This method allows to analyze  the new object and its region of competence to decide whether or not it is located in an indecision region (the region of competence with samples from different classes). The method proposed in~\cite{cruz2019fire} enhances the previous algorithm and reduces the overlap of classes in the validation set. It also defines the region of competence using an equal number of samples from each class. The combination of the static and dynamic ensemble selection was proposed in~\cite{nguyen2020ensemble}. In this ensemble selection method a base classifier is selected to predict a test sample if the confidence in its prediction is higher than its credibility threshold. The credibility thresholds of the base classifiers are found by minimizing the empirical $0–1$ loss on the entire training observations. The optimization-based approach to ensemble pruning is proposed in~\cite{bian2019ensemble}, where from an information entropy perspective an objective function used in the selection process is proposed. This function takes diversity and accuracy into consideration both implicitly and simultaneously.

The problem of the ensemble selection is considered, inter alia, from the point of view of the difficulty of data sets being classified~\cite{brun2018framework}. An example of the difficult data are imbalanced datasets in which there is a large disproportion in the number of objects from the particular class labels. As shown in~\cite{roy2018study}, the dynamic ensemble selection, due to the nearest neighbors analysis of the newly classified object, leads to better classification performance than the static ensemble pruning. The subject of classifiers selection for imbalanced data is up-to-date, and the existing methods are modified to obtain higher values of classification performance measures~\cite{junior2020novel}.

\section{The proposed framework}\label{sec:mehod}

We propose CS framework with a structure consisting of the following steps: (1) learning base classifiers, (2) relearning base classifiers, (3) feature generation based on learned and relearned base classifiers, (4) learning second-level base classifier based on a new vector of the features and (5) selection base classifiers based on second-level classification result. 

\subsection{Generation of diverse base classifiers}

Generating a diverse set of base classifiers is one of the essential points in creating an MCS. We propose that the set of homogeneous base classifiers should be generated using the bagging method. We denote as $D_{k}$ the training dataset of $\Psi^B_k$ base classifier, where $k \in \{1,...,K\}$, $K$ is the number of base classifiers.

\subsection{Relearning base classifiers}

Suppose we consider the problem of binary classification. Then each object can belong to one of the two class labels $\{-1,1\}$. So we define two new training sets $DS^{(-1)}$ and $DS^{(1)}$. These datasets contain objects from $D_k$ and new objects (or objects set) with arbitrarily assigned class labels. For example, in the dataset $DS^{(-1)}$, a new object or set of objects has an arbitrarily assigned class label $-1$. This dataset is used to relearning base classifiers separately, i.e. base classifiers labeled as $\Psi^{R(-1)}_k$ and $\Psi^{R(1)}_k$. 

\subsection{Feature generation based on learned and relearned base classifiers}

The object $x_i$ is represented in $d$ dimensions feature space as an vector $x_i=\left(x^1_i,...,x^d_i\right)$.
Based on learned base classifiers $\Psi^B_k$ and relearned base classifiers $\Psi^{R(-1)}_k$ and $\Psi^{R(1)}_k$ we can generate the new features for the new object or set of objects. We propose to add new feature to the object vector as the following features:
\begin{itemize}
\item the score returned by base classifier $\Psi_k^B(x_i)$,
\item the difference between score function returned by base and relearned classifier for added class label $-1$:  $\sigma^{(-1)}_{i,k}=|\Psi_k^B(x_i)-\Psi^{R(-1)}_k(x_i)|$,
\item the difference between score function returned by base and relearned classifier for added class label $1$:  $\sigma^{(1)}_{i,k}=|\Psi_k^B(x_i)-\Psi^{R(1)}_k(x_i)|$,
\item the information whether the object $x_i$ is correctly classified by base classifier $I\left(\Psi_k^B(x_i),\omega_{x_i}\right)$, 
\end{itemize}
where $I\left(\Psi_k^B(x_i),\omega_{x_i}\right)$ return $0$ when object $x_i$ is incorrect classified by base classifier $\Psi_k^B(x_i)$ and $1$ in the case of correctly classification.

    \begin{figure}[!htb]
        \centering
        \includegraphics[width=10cm]{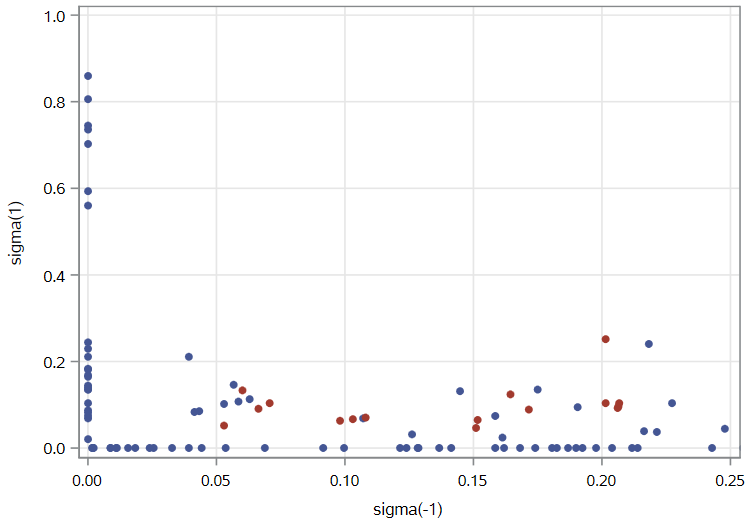}
        \caption{Correctly (blue points) and incorrect (red points) classified object in $\sigma^{(-1)}_{i,k}$,  $\sigma^{(1)}_{i,k}$ feature space.}
        \label{fig:c2}
    \end{figure}

The visualization of the two new features ($\sigma^{(-1)}_{i,k}$,  $\sigma^{(1)}_{i,k}$ and $I\left(\Psi_k^B(x_i),\omega_{x_i}\right)$) for one base classifier $\Psi^B_k$ and one set of object $x_i$ form validation dataset (synthetic Hygleman dataset used in experiments) is shown in Figure~\ref{fig:c2}. The red points represent  incorrect classified object by base classifier $\Psi^B_k$, while blue points represent correct classified object by base classifier $\Psi^B_k$. As it is easy to see objects incorrectly classified in the space defined by the features $\sigma^{(-1)}_{i,k}$,  $\sigma^{(1)}_{i,k}$ are located close to each other. This observation is the basis for the next step in which generated features are used to CS process.

\subsection{Learning second-level base classifier based on new vector of the features and}

In this step, we propose to learn base classifiers using the dataset using in the previous step. It means that second level base classifiers used also features generated in the previous step. The learning process takes into account only objects for which new features have been generated. The aim of learning at the second level is to build a model that will determine for the newly classified object $x_0$ and its additional features whether the base classifier is removed from the pool.

\subsection{Selection base classifiers based on second-level classification result}

Results from the previous step are used in CS process. The $x_0$ object with an arbitrarily assigned class label is used to retrain the base classifiers. Next, the new features of object $x_0$ are generated according to the previously described procedure. Afterward, the object $x_0$ is classified using the second level base classifiers. The classifiers obtained from these classifiers determine whether the base classifier from the first level is selected for the final classifiers pool. The entire proposed CS framework for the binary problem is presented in the Algorithm~\ref{RBalg1}.

 \begin{algorithm}[!ht]
\small
\KwIn{Dataset $D$, new object $x_0$, number of base classifier $K$}
\KwOut{The ensemble decision after the relearning ensemble selection based on new generated features}

Split $D$ into: training $D_{tr}$ dataset and the validation dataset $D_{va}$

Split $D_{tr}$ into $K$ folds $D_1,...,D_{K}$ by bagging procedure  

$ \forall_{D_k} \forall_{x_i \in D_{va}}$ add $x_i$ with class label $=-1$ to $D_k$ -- new fold $DS^{(-1)}_{i, k}$ 

$\forall_{D_k} \forall_{x_i \in D_{va}}$ add $x_i$ with class label $=1$ to $D_k$ -- new fold $DS^{(1)}_{i, k}$ 

Train base classifier $\Psi_k^B$ using $D_k$
  
Train the relearned base classifiers for class label $-1$, $\Psi^{R(-1)}_k$ using $DS^{(-1)}_{i, k}$  

Train the relearned base classifiers for class label $1$, $\Psi^{R(1)}_i$ using $DS^{(1)}_{i, k}$

$\forall_{D_k} \forall_{x_i \in D_{va}}$ calculate $\sigma^{(-1)}_{i,k}=|\Psi_k^B(x_i)-\Psi^{R(-1)}_k(x_i)|$ and 
$\sigma^{(1)}_{i,k}=|\Psi_k^B(x_i)-\Psi^{R(1)}_k(x_i)|$ 

Create $K$ new learning datasets $D^R_{k}:$ $$\forall_{x_i \in D_{va}} \quad x^R_{i,k}=\left(x^1_i,...,x^d_i,\Psi_k^B(x_i), \sigma^{(-1)}_{i,k}, \sigma^{(1)}_{i,k}, I(\Psi_k^B(x_i),\omega_{x_i})\right)$$

Add $x_0$ with class label $=-1$ to $D_k$ -- new fold $DS^{(-1)}_{0,k}$ 

Add $x_0$ with class label $=1$ to $D_k$ -- new fold $DS^{(1)}_{0,k}$ 

Crate new features for $x_0$ using trained base classifiers $\Psi_k^B$ and trained the learned base classifiers on datasets $DS^{(-1)}_{0,k}$ and $DS^{(1)}_{0,k}$: $$x_{0,k}=\left(x^1_0,...,x^d_0,\Psi_k^B(x_0), \sigma^{(-1)}_{0,k}, \sigma^{(1)}_{0,k}\right)$$

Train second level base classifiers $\Psi_k^{BSL}$ using $D^R_{k}$

If $\Psi_k^{BSL}(x_0)=0$, then $\Psi^{RES}_i(x_0)=\Psi_i^B(x_0)$ else $\Psi^{RES}_i(x_0)=0$

The ensemble decision after the relearned ensemble selection:
$$\Psi^{RES}(x_0)=sign\left(\sum_{i=1}^{K} \Psi^{RES}_i(x_0)\right),$$
where $\Psi^{RES}_i(x_0)$ return value in the range $(-1, 0)$ for predicted class label $-1$ and $(0, 1)$ for predicted class label $1$.

\caption{Relearning ensemble selection algorithm based on new generated features -- for the binary problem}
\label{RBalg1}
\end{algorithm}

\section{Experiments}\label{sec:experimental-setup}

\subsection{Experimental setup} 

The implementation of SVM from SAS 9.4 Software was used to conduct the experiments. In experiments, we took into account the radial basis function as the kernel function. The regularization parameter $C$ was searched in the set $C \in  \{0.001, 0.01, 0.1, 1, 10, 100\}$ using grid search procedure~\cite{scholkopf2018learning}.
This parameter was search separately for train fort and second level base classifiers. In the experiments, $5x2$ cross-validation method has been used. As a validation dataset we use a learning dataset, which means that we use the resubstitution method for new features generation.

    \begin{table}[!htb]
        \centering
        \small
        \caption{Descriptions of datasets used in experiments (name with abbreviation, number of instances, number of features, imbalance ratio).}
        \label{tab:datasets}
        \begin{tabular}{|c|c|c|c|}
            \hline
            Dataset                               & \#inst & \#f & Imb   \\ \hline 
            Breast Cancer -- original (Cancer)          & 699    & 9   & 1.9   \\ \hline
            Liver Disorders (Bupa)            & 345    & 6   & 1.4   \\ \hline
            Pima (Pima)                             & 768    & 8   & 1.9   \\ \hline
            Synthetic dataset (Syn)                             & 400    & 2   & 1   \\ \hline
            \end{tabular}
    \end{table}

A performance classification metric such as the area under the curve (AUC), the G-mean (G), the F-1 score (F-1) and the Matthews correlation coefficient (MCC) have been used. As a reference ensemble of classifiers, we use majority voting ($\Psi^{MV}$) and sum rule without selection ($\Psi^{SUM}$).

    \begin{figure}[!htb]
        \centering
        \includegraphics[width=10cm]{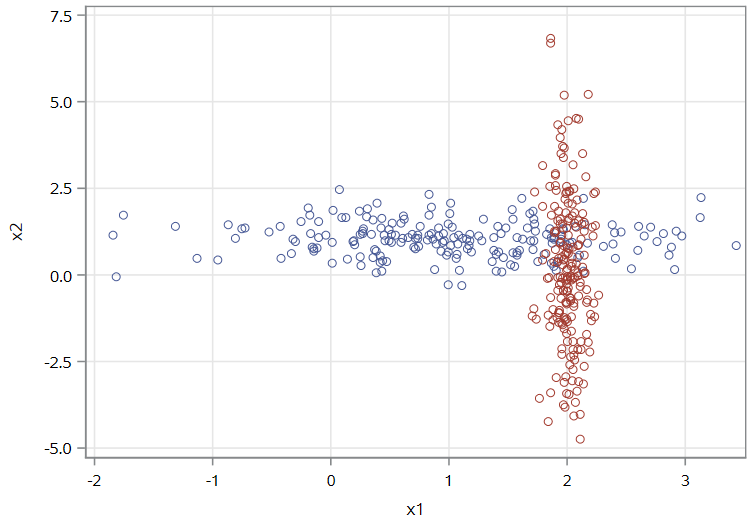}
        \caption{Synthetic datasest used in the experiment.}
        \label{fig:sds}
    \end{figure}

Three real datasets from UCI~\cite{uci} repository and one synthetic dataset were used in the experiments. The synthetic dataset used in the experiment is presented in Figure~\ref{fig:sds}. The datasets that were used to validate the algorithms are presented in Table~\ref{tab:datasets}.
The number of instances, features, and imbalance ratio were included in the description.


\subsection{Results}

The experiments were conducted in order to compare the classification performance metrics of the proposed relearning ensemble selection based on new generated features algorithm $\Psi^{RES}$ with referential ensemble techniques: majority voting ($\Psi^{MV}$) and sum rule without selection ($\Psi^{SUM}$). The results of the experiment for four classification performance metrics are presented in Table~\ref{tab:result}. The bold letters indicate the best results for each performance metric and each database separately. 
    
\begin{table}[!htb]
\centering
 \small
 \caption{Result of classification.}
  \label{tab:result}
\begin{tabular}{c|c|c|c|c|c|}
\cline{3-6}
\multicolumn{2}{l|}{}                               & AUC  & G & F-1   & MCC   \\ \hline
\multicolumn{1}{|c|}{\multirow{3}{*}{Cancer}} & $\Psi^{SUM}$ & 0.942 & 0.940  & 0.941 & 0.806 \\ \cline{2-6} 
\multicolumn{1}{|c|}{}                        & $\Psi^{RES}$ & \bf{0.944} & \bf{0.943}  & \bf{0.944} & \bf{0.812} \\ \cline{2-6} 
\multicolumn{1}{|c|}{}                        & $\Psi^{MV}$ & 0.942 & 0.940  & 0.941 & 0.806 \\ \hline
\multicolumn{1}{|c|}{\multirow{3}{*}{Bupa}}   & $\Psi^{SUM}$ & 0.537 & 0.487  & 0.674 & 0.083 \\ \cline{2-6} 
\multicolumn{1}{|c|}{}                        & $\Psi^{RES}$ & \bf{0.543} & 0.494  & \bf{0.678} & \bf{0.096} \\ \cline{2-6} 
\multicolumn{1}{|c|}{}                        & $\Psi_{MV}$ & 0.540 & \bf{0.495}  & 0.673 & 0.087 \\ \hline
\multicolumn{1}{|c|}{\multirow{3}{*}{Pima}}   & $\Psi^{SUM}$ & 0.715 & 0.693  & 0.832 & 0.465 \\ \cline{2-6} 
\multicolumn{1}{|c|}{}                        & $\Psi^{RES}$ & 0.721 & \bf{0.703}  & 0.833 & 0.473 \\ \cline{2-6} 
\multicolumn{1}{|c|}{}                        & $\Psi^{MV}$ & \bf{0.722} & \bf{0.703}  & \bf{0.835} & \bf{0.478} \\ \hline
\multicolumn{1}{|c|}{\multirow{3}{*}{Syn}}    & $\Psi^{SUM}$ & 0.845 & 0.844  & 0.840 & 0.691 \\ \cline{2-6} 
\multicolumn{1}{|c|}{}                        & $\Psi^{RES}$ & \bf{0.875} & \bf{0.875}  & \bf{0.872} & \bf{0.751} \\ \cline{2-6} 
\multicolumn{1}{|c|}{}                        & $\Psi^{MV}$ & 0.840 & 0.839  & 0.834 & 0.682 \\ \hline
\end{tabular}
\end{table}

\section{Discussion}~\label{sec:Conclusions}

It should be noted that the proposed algorithm may improve the quality of the classification compared to the reference methods. In particular, for the synthetic set, the improvement is visible for all analyzed metrics and exceeds the value of $3\%$. In the case of real datasets, the improvement of the value of the metrics is not so significant, but the method without selection $\Psi^{SUM}$ is always worse than the proposed algorithm $\Psi^{RES}$. The conducted experiments concern one test scenario in which the SVM base classifiers were learned using the bagging procedure.

We treat our research as a preliminary study. The directions of further research include:
\begin{itemize}
\item evaluation of larger groups base classifiers,
\item evaluation of larger number of objects in the retraining stage,
\item development of a new features dedicated to semi-supervised problem.
\item development of a new feature dedicated to the problem of decomposition of a multi-class task that eliminate the problem of incompetent classifier,
\item development of a new features dedicated to imbalanced dataset problem.
\end{itemize}

\section{Conclusions}
This paper presents a new approach to the CS process. In the proposal described in the article, the selection process of the new generated feature obtained after relearning base classifiers is used. Additionally, the second-level base classifier is learning based on new generated features. The classification models obtained after second-level learning are used to select base classifiers (from the first-level).

The experimental results show that the proposed method can obtain better classification results than the reference methods. Such results were obtained for four different performance classification measures.

Presented in the paper, research on relearning ensemble selection is a preliminary study. Based on the obtained promising results, future research will focus, among other things, on the development of new features dedicated to semi-supervised, multi-class, and imbalanced dataset problems.

\bibliographystyle{plain} 
\bibliography{bibliography}

\end{document}